\documentclass[preprint,12pt]{elsarticle}

\usepackage{amssymb}
\usepackage{color}
\usepackage{fontawesome}
\usepackage[dvipsnames]{xcolor}
\usepackage{moreverb,url}
\usepackage{graphicx}
\usepackage{dcolumn}
\usepackage{multirow}
\usepackage{bm}
\usepackage[utf8]{inputenc}
\usepackage{amsmath}

\usepackage[colorlinks,bookmarksopen,bookmarksnumbered,citecolor=red,urlcolor=red]{hyperref}

\journal{Sustainable Cities and Society}

\begin{document}

\begin{frontmatter}



\title{Predicting the traffic flux in the city of Valencia with Deep Learning}


\author[UPV]{Miguel G. Folgado}
\ead{migarfol@upvnet.upv.es}
\author[UV,Sussex]{Ver\'onica Sanz}
\ead{veronica.sanz@uv.es}
\author[UI1,AI]{Johannes Hirn}
\ead{johanneskaspar.hirn@ui1.es}
\author[UPV]{Edgar G. Lorenzo}
\ead{edlosae@etsiamn.upv.es}
\author[UPV]{Javier F. Urchueguía}
\ead{jfurchueguia@fis.upv.es}

\affiliation[UPV]{organization={ICTvsCC research group, Instituto Universitario de Tecnolog\unexpanded{í}as de la Informaci\unexpanded{ó}n y Comunicaciones (ITACA)\unexpanded{,} Universidad Polit\unexpanded{é}cnica de Valencia},
            addressline={Edificio 8G Ciudad Polit\unexpanded{é}cnica de la Innovaci\unexpanded{ó}n\unexpanded{,} Cam\unexpanded{í} de Vera\unexpanded{,} s\unexpanded{/}n},
            city={Valencia},
            postcode={46022}, 
            country={Spain}}

\affiliation[UV]{organization={Instituto de F\unexpanded{í}sica Corpuscular (IFIC)\unexpanded{,} Universidad de Valencia-CSIC},
            addressline={Carrer del Catedr\unexpanded{á}tic Jos\unexpanded{é} Beltr\unexpanded{á}n Martinez, 2},
            city={Paterna},
            postcode={46980},
            country={Spain}}

\affiliation[Sussex]{organization={School of Mathematical and Physical Sciences, University of Sussex},
            addressline={Falmer Campus},
            city={Brighton},
            postcode={BN1 9RH}, 
            country={United Kingdom}}
            
\affiliation[UI1]{organization={Faculdad de Ciencias y Tecnolog\unexpanded{í}a, Universidad Isabel I de Castilla},
            addressline={Calle Fern\unexpanded{á}n Gonz\unexpanded{á}lez, 76},
            city={Burgos},
            postcode={09003},
            country={Spain}}

\affiliation[AI]{organization={AI Superior GmbH},addressline={Robert-Bosch-Str. 7,
64293 Darmstadt}, country={Germany}}            
            
\begin{abstract}
Traffic congestion is a major urban issue due to its adverse effects on health and the environment, so much so that reducing it has become a priority for urban decision-makers. In this work, we investigate whether a high amount of data on traffic flow throughout a city and the knowledge of the road city network allows an Artificial Intelligence to predict the traffic flux far enough in advance in order to enable emission reduction measures such as those linked to the Low Emission Zone policies. To build a predictive model, we use the city of Valencia traffic sensor system, one of the densest in the world, with nearly 3500 sensors distributed throughout the city. In this work we train and characterize an LSTM (Long Short-Term Memory) Neural Network to predict temporal patterns of traffic in the city using historical data from the years 2016 and 2017. We show that the LSTM is capable of predicting future evolution of the traffic flux in real-time, by extracting  patterns out of the measured data. 
\end{abstract}


\end{frontmatter}

\section{Introduction}

Road traffic causes one-fifth of the total greenhouse gas (GHG) emissions in the European Union~\cite{EC,EEA,EEA2}. Specifically, for the city of Valencia, GHG emissions from road transport represent $60\%$ of its total GHG emissions~\cite{GHG}. Additionally, vehicle traffic is one the main contributors to air pollution in the urban environment~\cite{art5}. Road vehicles emit pollutants in various ways --- exhaust, abrasion and resuspension--- the combination of which has an important impact on air quality \cite{art5}. 

Exposure to elevated concentrations of Dioxide Nitrogen (NO2) and Particulate matter (PM) are the main risk factors for adverse health effects and premature deaths worldwide~\cite{art1,WHO}. For example, exposure to PM (especially the  fine fraction) is correlated with the outbreaks of allergy aggravation, respiratory, cardiovascular and even cerebrovascular diseases~\cite{art2,art3,art4}. Additionally, exposure to elevated concentrations of NO2 in the air is linked to a range of respiratory diseases such as bronchoconstrictions, increased bronchial reactivity, airway inflammation and decreases in immune defenses leading to increased susceptibility to respiratory tract infection~\cite{UK}. This makes traffic congestion a major urban problem~\cite{10.7864/j.ctt1vjqprt,art6}, and  reducing it has become a priority for urban decision-makers.

Some authors~\cite{RePEc:mtp:titles:0262012197} demonstrate the necessity of applying a microscopic approach to identify adequate mitigating actions for traffic congestion. In this view, high spatial and temporal resolutions are used to study traffic flows and patterns in the different traffic hotspots of the city.

The main objective of our work is to develop a Neural Network algorithm to identify spatial and temporal patterns (the visible ones and the not so obvious) and to predict traffic with high spatial and temporal resolution.

\section{Related work}

Several authors have developed  models to identify the microscopic problems of  traffic flux: examples can be found in Refs.~\cite{art7,PhysRevE.51.1035,RePEc:inm:oropre:v:9:y:1961:i:4:p:545-567}. These models are based on mathematical descriptions of various aspects of mobility, accounting for velocities and density of vehicles for instance. In this work, we present a different type of analysis, based on using  a large amount of traffic flux data  and the knowledge of the road city network. We then train an  Artificial Intelligence to predict the traffic flux sometime far ahead in the future, potentially allowing policy-makers to take emission reduction measures. 

The spectrum of traffic features that one can forecast is broad, for example, traffic speed~\cite{a8}, travel time~\cite{a36}, as well as traffic flux~\cite{a37}. In order to reduce emissions, we are primarily interested in the predictions of  traffic flux: this variable will be the target of the study.

Several authors have studied how Neural Networks can predict different features of a traffic network. Some examples are based on long short-term memory (LSTMs)~\cite{2018arXiv181104745M,Song2016DeepTransportPA}, simpler gated recurrent unit (GRU)~\cite{2014arXiv1406.1078C}, deep belief networks (DBN)~\cite{Huang2014DeepAF,a12}, stacked autoencoders ~\cite{a2} and generative adversarial networks \cite{2018arXiv180103818L}. More recent studies have used graph neural networks (GraphNN) to tackle this problem~\cite{2016arXiv160902907K,2013arXiv1312.6203B,2015arXiv151102136A,2016arXiv160601166V}. Also, some authors have studied the potential of the combination of different kinds of Neural Networks, for example see Ref.~\cite{2018arXiv180207007C}. 

\section{Descriptive study}
\subsection{Data-set description}
\label{sect:dataframe}

This study is based on data from induction loops used to monitor traffic in the city of Valencia, composed of 3500 sensors distributed in 1500 road segments. These induction loops provide data of total traffic intensity (number of vehicles) and velocity at a given induction loop, within a prescribed time window.  The data obtained from the induction loops is registered at the Smart City València infrastructure~\cite{smartcity} and automatically transferred to the Universitat Politècnica de València. 

Historical data from 2016 and 2017 for each of these road segments, has been used in this work. To illustrate the practical use of the data in a future real-time pipeline, we used a  time resolution of one hour in the predictive analysis, although the sensors  can provide data with a 5-minute frequency. 

 The data is first processed to remove low-quality measurements, along with statistical calculations to obtain average values and deviations. In particular, before averaging over each hour, outliers are removed from the 5-minute frequency data to keep values in the range:
\begin{equation}
    m-3\sigma \leq v \leq m+3\sigma \ ,
\end{equation}
where $m$ is the road segment flux average over that hour, $\sigma$ is the typical deviation and $v$ the value of the evaluated data.

\subsection{Geographical distribution of the sensorized road segments}

\label{sec:sample1}

In our analysis, we define the sensorized road sections as the sections where the traffic flux is completely determined by our sensors. In this sense, each sensor measures the traffic between two intersections, and its exact location along the road segment linking those two intersections is irrelevant in our analysis, up to possible vehicles parking along that segment.

\begin{figure}[h]
\centering
\includegraphics[width = 1\textwidth]{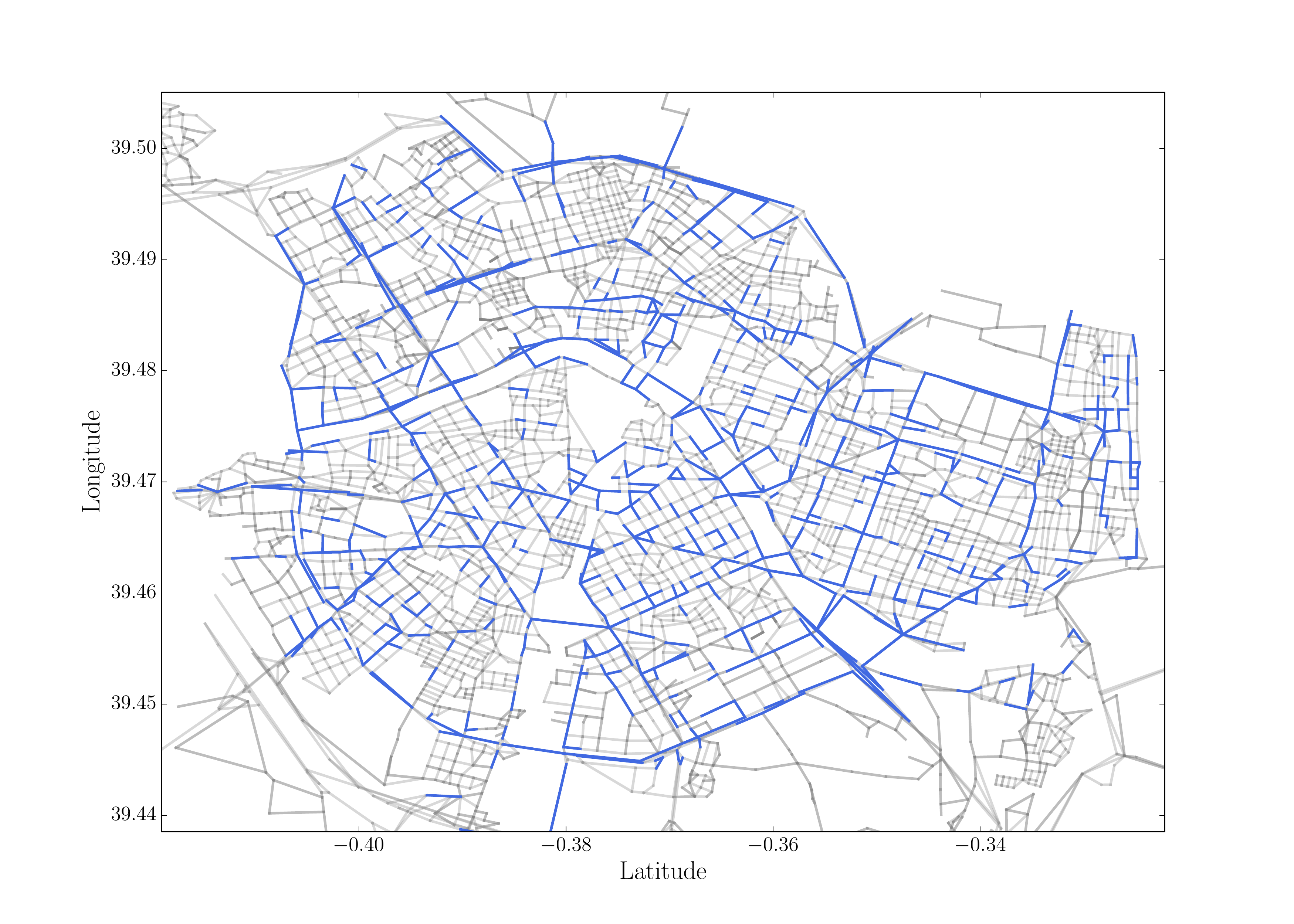}
\caption{Map of the distribution of sensors in the road network of Valencia city. In grey we show the different road segments of Valencia's traffic network. The blue lines represent the road segments for which we have flux data, i.e., the sensorized road segments.}
\label{fig:sensores}
\end{figure}

In Fig.~\ref{fig:sensores} we show the different sensorized road sections (blue segments) for which we have obtained the traffic flux. The initial granularity of measurements was five minutes. However, for the purposes of a predictive analysis such time-resolution is unnecessary, as realistic  decisions making and implementation will not be as fast and we will focus on predicting  flux one to  two hours ahead. 

As can be seen in Fig.~\ref{fig:sensores}, these sections represent only a small part of the complete road network of Valencia city. To be specific, the area being studied consists of 8595 road sections, of which less than $20\%$ are sensorized.

\subsection{Traffic flux}

As mentioned in Sec.~\ref{sect:dataframe}, we have data on flux and velocities from the sensorized roads of Valencia city. In our analysis, we are going to focus on the number of  vehicles going through the monitored segments. We will define  a measurement of traffic  flux as the number of vehicles per unit of time. In our data, the time resolution will be one hour.  

With this definition of flux, we move to conduct a descriptive analysis of the evolution of the flux with time. We will focus on {\it 1.)} the total traffic flux in all segments, {\it 2.)} the daily patterns for specific sections in the city (main access roads), {\it 3.)} the differences between incoming and outgoing flux, and {\it 4.)} rush hour traffic.

\subsubsection{Analysis of the total traffic flux}

\label{sec:weekly_flux}

We start analysing the data by looking at a global quantity, the total flux measured  in all the sensorized segments. One would expect to see a weekly pattern, with quieter activity in the weekends and an increased flux during working  days. 

\begin{figure}[h!]

\centering

\includegraphics[width = 1.0\textwidth]{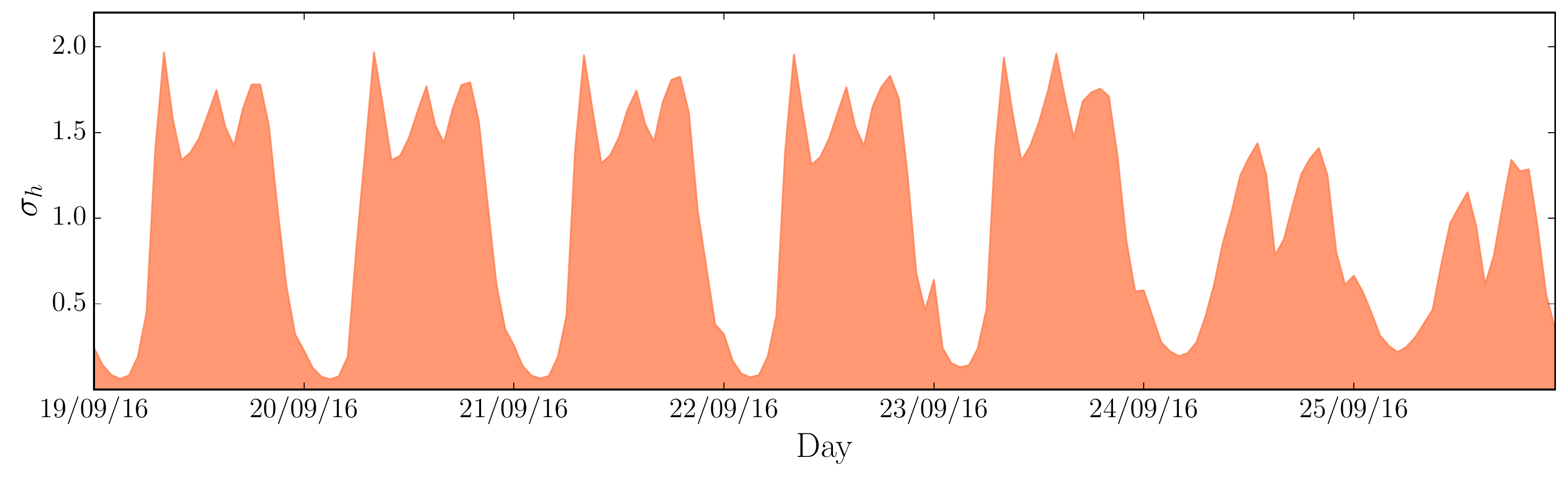}

\caption{The orange-shaded represents the deviation $\sigma_h = N_h / \bar{N}$ from the mean weekly flux, in the studied week (from Monday, September 19, 2016 till Sunday, September 25, 2016).}

\label{fig:weekly_flux}

\end{figure}

This is indeed the pattern found in the data, as  shown in Fig.~\ref{fig:weekly_flux}. In this Figure  we plot the number of vehicles per hour respect to its mean over the week, $\sigma_h=N_h/\tilde N$. Values of $\sigma_h$ below (above) 1 represent periods that are quieter (busier) than average.

On the x-axis we chose a typical week, corresponding to September 2016, from Monday to Sunday.  

As we can see in the plot, all days have a similar overall behavior, with quiet times during the night. But one can see marked differences across the week.

Each working day exhibits three peaks: times where people would travel to/from work or school, and a third peak around midday when people travel for lunchtime, typically back home.  

The pattern of Friday is noticeably different from the rest of the working days, with an increased lunch-hour peak and a reduced afternoon peak, as  workers may leave the office at lunchtime and not come back on Friday afternoon. 

The pattern changes again during the weekend days, Saturday and Sunday. These two days we only observe two clear peaks, possibly related to midday shopping and evening social outings. We will discuss these effects in more detail  in the next two subsections.

\subsubsection{Analysis of the daily patterns in the main roads}

The analysis of Sec.~\ref{sec:weekly_flux} clearly indicates different patterns in the flux during the days in a standard week. Now we are going to move closer and analyze the evolution during the day, already visible in  Fig.~\ref{fig:weekly_flux}.

\begin{figure}[h!]

\centering

\includegraphics[width = 0.99\textwidth]{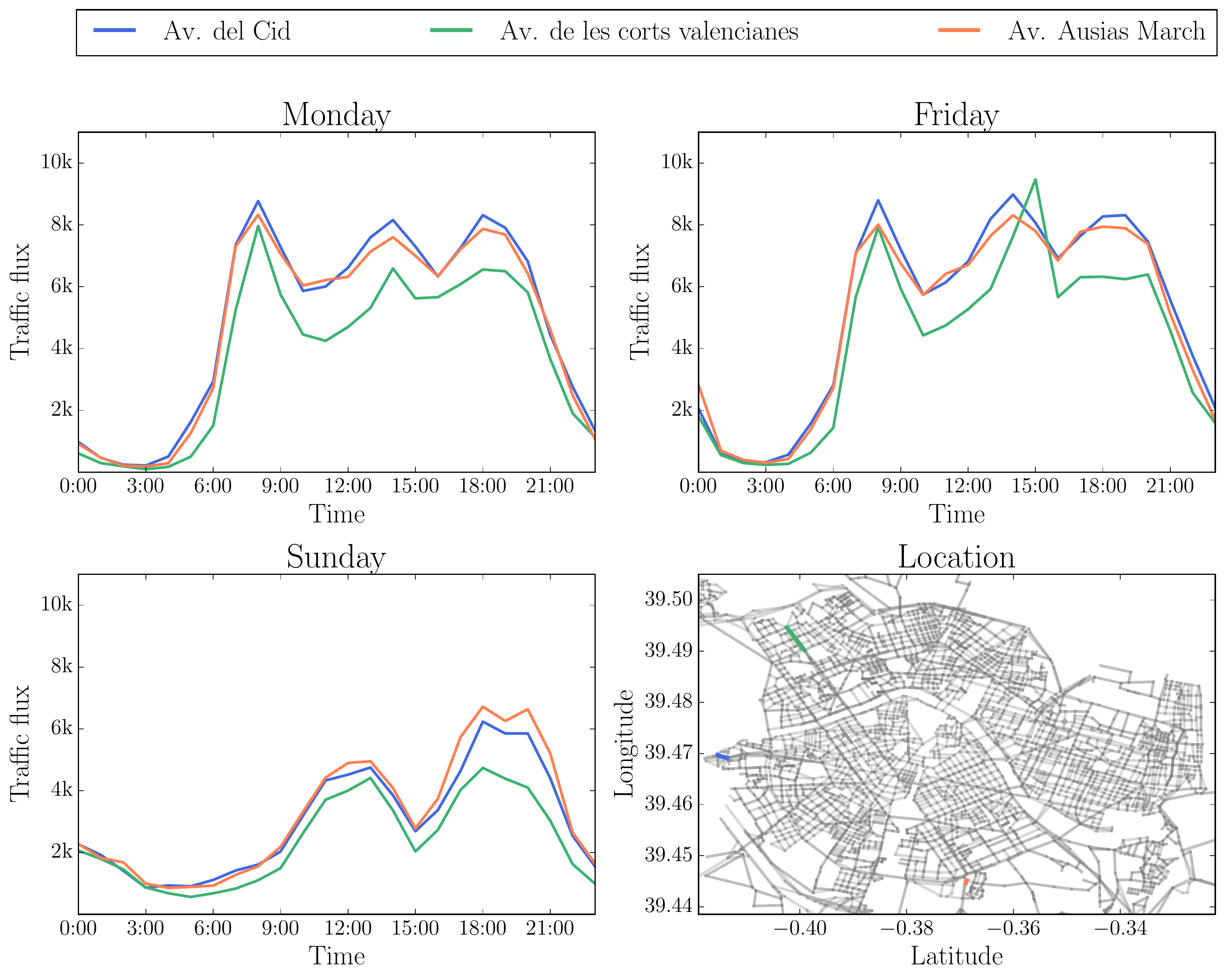}

\caption{Analysis of the traffic flux in the most traffic-heavy sensorized road sections. In the four panels we show In the top-left panel the traffic flux evolution of 19/09/16 (Monday); in the top-right panel the traffic flux evolution of 23/09/16 (Friday); in the bottom-left panel the evolution of 25/08/16 (Sunday); and, finally, in the bottom-right panel we show the analysed road segments.}

\label{fig:patrones}

\end{figure}

In the previous section we showed that the weekly patterns were clustered around three types of behaviour: Monday through Thursday, weekends  and Friday.  To analyse this pattern in more detail, we will focus on the main exit and entry roads in the city of Valencia: {\color{blue}Av del Cid} to the West,  {\color{teal} Av de les corts valencianes} to the North and { \color{orange} Av Ausias March} to the South\footnote{Valencia is bordered by the sea on the East side.}. We have determined in our analysis that  these three segments represent the busiest roads in the city. The segments are located at the main access points to the city. The location of these avenues is shown in the lower right panel of Fig.~\ref{fig:patrones}.

The blue, green and orange lines in the graphs represent, respectively, segments located in {\color{blue} Av. del Cid}, {\color{teal}  Av. de les cortes valencianes} and {\color{orange} Av. Ausias March}. The top-left panel corresponds to the flux on a weekday (Monday 19/09/16), with a pattern that will be replicated during the rest of the week until Friday. 
We observe three characteristic peaks in the three segments, located around 8 AM, 2 PM and 7 PM.

If we focus on the early morning rush hour, we can distinguish different times for the start  of the working day. There is a sharp increase of traffic from  6 AM, while the maximum is reached around 8 AM, related to the usual office working hours which start at 9 AM.  

The next peak, located around 1 PM and likely related to  lunchtime, is less acute
in {\color{teal} Av de les cortes valencianes}, than in  the other two sections. This could be related to different lunch patterns, where workers coming from the North of the city tend to eat near their workplace, but workers from the South of the city keep the traditional custom of lunching at home. Another factor at play in this midday peak could be the work modality called {\it jornada continua}, where workers start early (8 AM) and leave the workplace at around 3 PM, with no lunch break. This modality is often found in public offices.  

Both effects (lunch and {\it jornada continua}) seem to be at play, as we will discuss in the next sub-section. There we will plot the incoming and outgoing traffic, Fig.~\ref{fig:entrada_salida}. In this  plot we see how there is indeed an incoming flux after lunch-time but also a strong overall  outgoing peak around 3 PM.  
The third traffic daily peak is found in the region 6 to 8 PM, when many workers will finish their duties and go back home.

Let us now focus on Fridays. The typical pattern can be seen in the top right panel of Fig.~\ref{fig:patrones} and one can observe again some differences between the West and South access to Valencia, and the North access in  Av. de les cortes valencianes. On Fridays, the traffic flux of vehicles going through the North has a sharp peak around 3PM, whereas in the other two main roads the flux pattern is similar to the other working days. This may indicate that many workers in the North of the city have more working-hours flexibility and/or are public workers with {\it jornada continua}.

Finally, we can analyse a typical weekend day in the bottom left panel of Fig.~\ref{fig:patrones}. These days, the mobility is concentrated around midday and 7 PM, indicating that these movements may be related with entertainment, social outings and shopping.

\subsubsection{Incoming and outgoing traffic}

\begin{figure}[h!]

\centering

\includegraphics[width = 1.00\textwidth]{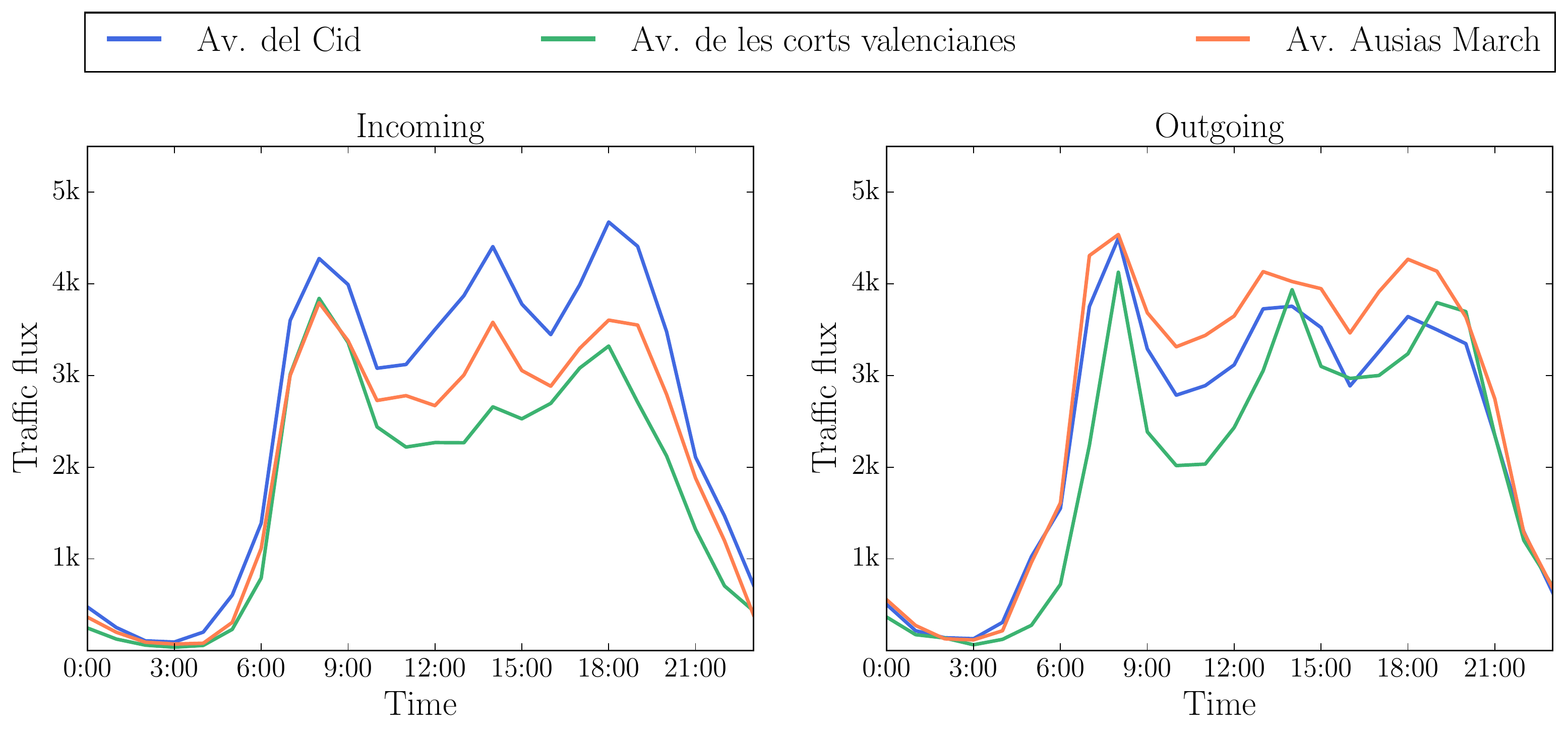}

\caption{Traffic flux in the three main accesses to the city for Monday, September 19, 2016. Left and right panels represents, respectively, the incoming and outgoing traffic.}

\label{fig:entrada_salida}

\end{figure}

In this section we will analyse the daily patterns, but now differentiating between incoming and outgoing flux. In Fig.~\ref{fig:entrada_salida} we show  the same Monday as in Fig.~\ref{fig:patrones}, but now plotting the flux coming to the city (incoming) and the flux going out of the city (outgoing).  

The three roads exhibit different behavior during the day. For example, the {\color{blue} Av. del Cid} is the main road for incoming traffic in the morning, however most morning exits occur through {\color{orange} Av. Ausias March}. This could  indicate that city dwellers who work outside the city travel to the industrial areas in the South  of the city.
It is also possible  that many drivers use GPS apps to find the quickest path to their work, and this could imply that the best incoming way  is probably  not the best outgoing.

\begin{figure}[h!]
\centering
\includegraphics[width = 1.00\textwidth]{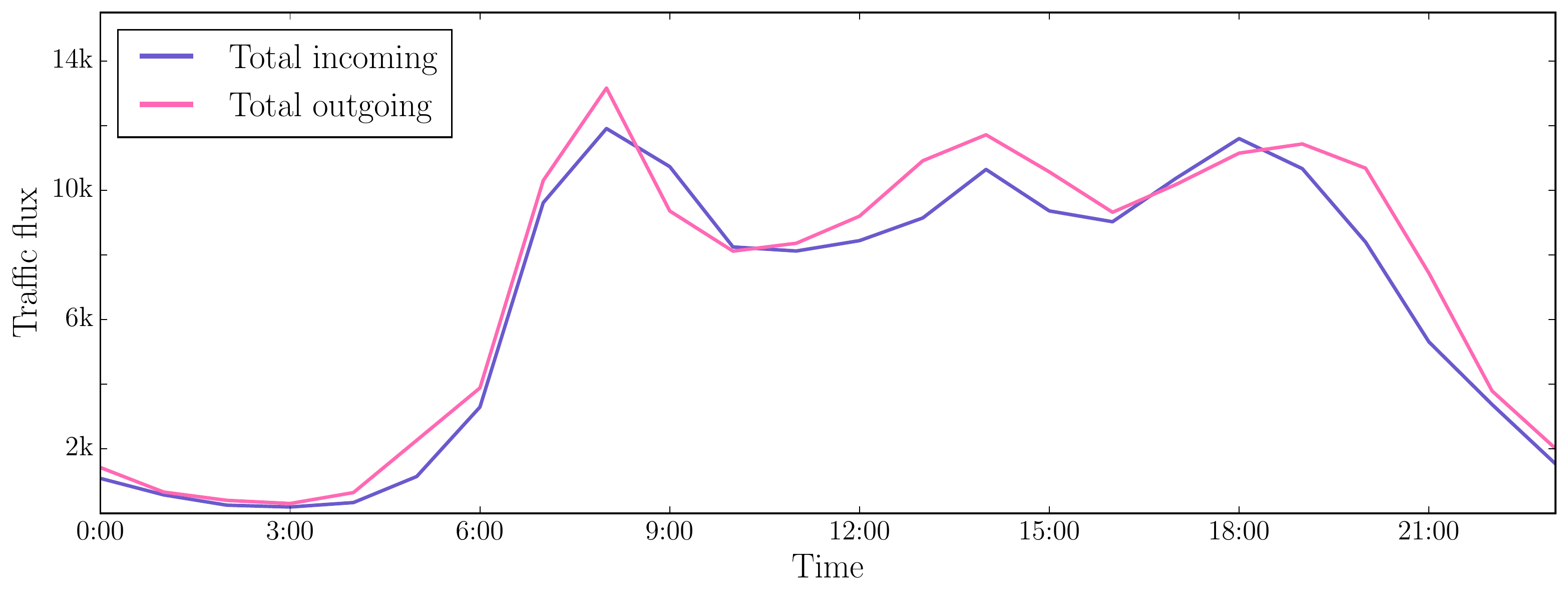}
\caption{Sum over the incoming and outgoing traffic on the three main roads for Monday, September 19, 2016.}
\label{fig:totales}
\end{figure}

It could also be the case that drivers choose smaller roads on one leg of return trip.  Nevertheless, the overall vehicle exchange on the three main road exchanges in the city of Valencia is close to zero, as  shown in Fig.~\ref{fig:totales}. 

\subsubsection{Rush hour}

To finish this descriptive analysis, we will  analyze the traffic flux during the two main rush hours in  a weekday, specifically at 8 AM and 7 PM.  

\begin{figure}[h!]

\centering

\includegraphics[width = 1.00\textwidth]{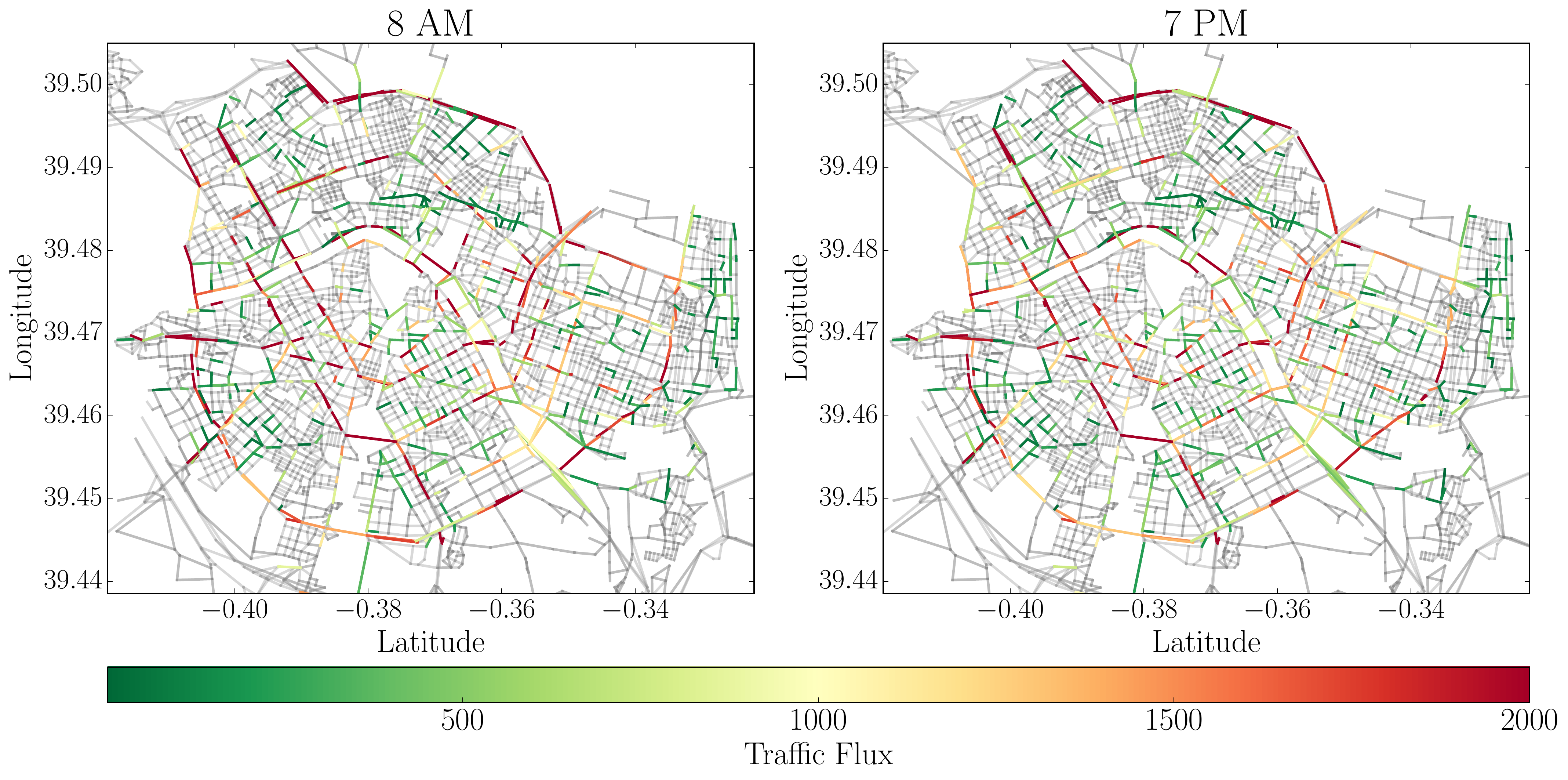}

\caption{Traffic flux per sensorized road in the rush hours of Monday, September 19, 2016. In the left and right panels we show the traffic flux at 8 AM and 7 PM, respectively. The color scale is clipped at 2000 to improve visibility.}

\label{fig:rush_hour}

\end{figure}

 For direct comparison with the previous plots, we show results from  Monday, September 19, 2016, although these results  are representative of the rush hours on all weekdays. As one can see in Fig.~\ref{fig:rush_hour}, the busiest roads are the main avenues and access points to the city.

\section{Predictive Analysis of the traffic}

\subsection{Analysis Description}

In the previous sections, we studied the different patterns of the traffic flux in Valencia city. As we have seen, there are several symmetries between different days of the week and different hours of the day. This indicates that modelling the traffic flux should be possible, as long as one can develop a model with enough complexity to account for these patterns.

The main idea of this work is to use a very expressive modelling tool based on Neural Networks, which could encapsulate the patterns we have identified, and others which may not be visible in a simple descriptive analysis.  The problem we would like   to focus on, evolution of the traffic in a city, has both spatial (traffic distribution on the road map) and time components (evolution with time).

Theoretically, one might expect that to achieve the best predictive power we should employ  Graph Neural Networks (GraphNN), as they are  ideal for these kind of problems.  Unfortunately, the current data does not allow for a successful GraphNN  analysis.  When we developed a predictive analysis using GraphNNs, the first problem we encountered  was related to the connectivity of the sensorized road segments. For a good graph description, it would be necessary to sensorize practically all the city road segments, and although we have explored different ways to overcome this issue, we find that the overall GraphNN  predictivity is worse than the method we will propose here.   

In addition to this problem, there is an issue of generalizability of the results. When developing a GraphNN analysis, and using some interpolations in segments which are not covered, the overall results ended up being very sensitive to the  geometry of the city. In this situation, it would not be easy to translate the GraphNN predictions to other cities, as it would be necessary to create a new graph for the new city and the interpolations we had made in the city of Valencia may not work for a different topology. 

Some of these shortcomings were already pointed out in previous studies~\cite{2018arXiv180207007C}, where they showed that including the map of the city in the Neural Network did not significantly increase the precision of the prediction. For these reasons, we chose to present our  analysis with a LSTM (Long Short-Term Memory)~\cite{10.1162/neco.1997.9.8.1735} Neural Network, instead of a GraphNN.

In the LSTM analysis, the inputs are time-ordered flux values corresponding to  measured segments labelled by an ID number. The LSTM will learn to predict the future evolution of the flux in these segments, given a past behaviour.

Note that the  spatial patterns of the city will be present in the flux data, but their locality will not be explicit. For example, the input to the LSTM  will not have any information of whether two ID segments are close to each other. Nevertheless, with enough data, the LSTM may  extract some of these patterns from the measured flux data.

First, we must define the data that we will use in the training of the Neural Network. In all of the results we will present, the LSTM is trained using the measured flux during the 366 days of the year 2016. Once trained on this data, we will test it on unseen data from 2017. In particular, we will predict the traffic flux in six days of February 2017 (more concretely, from Thursday 2 to Wednesday 8). Also, to test the precision of the LSTM in specific locations, we have studied the real and the predicted flux over the congested road segment located in {\color{blue} Av. del Cid}, one of the main arteries of Valencia city.

\begin{figure}[h!]

\centering

\includegraphics[width = 1.00\textwidth]{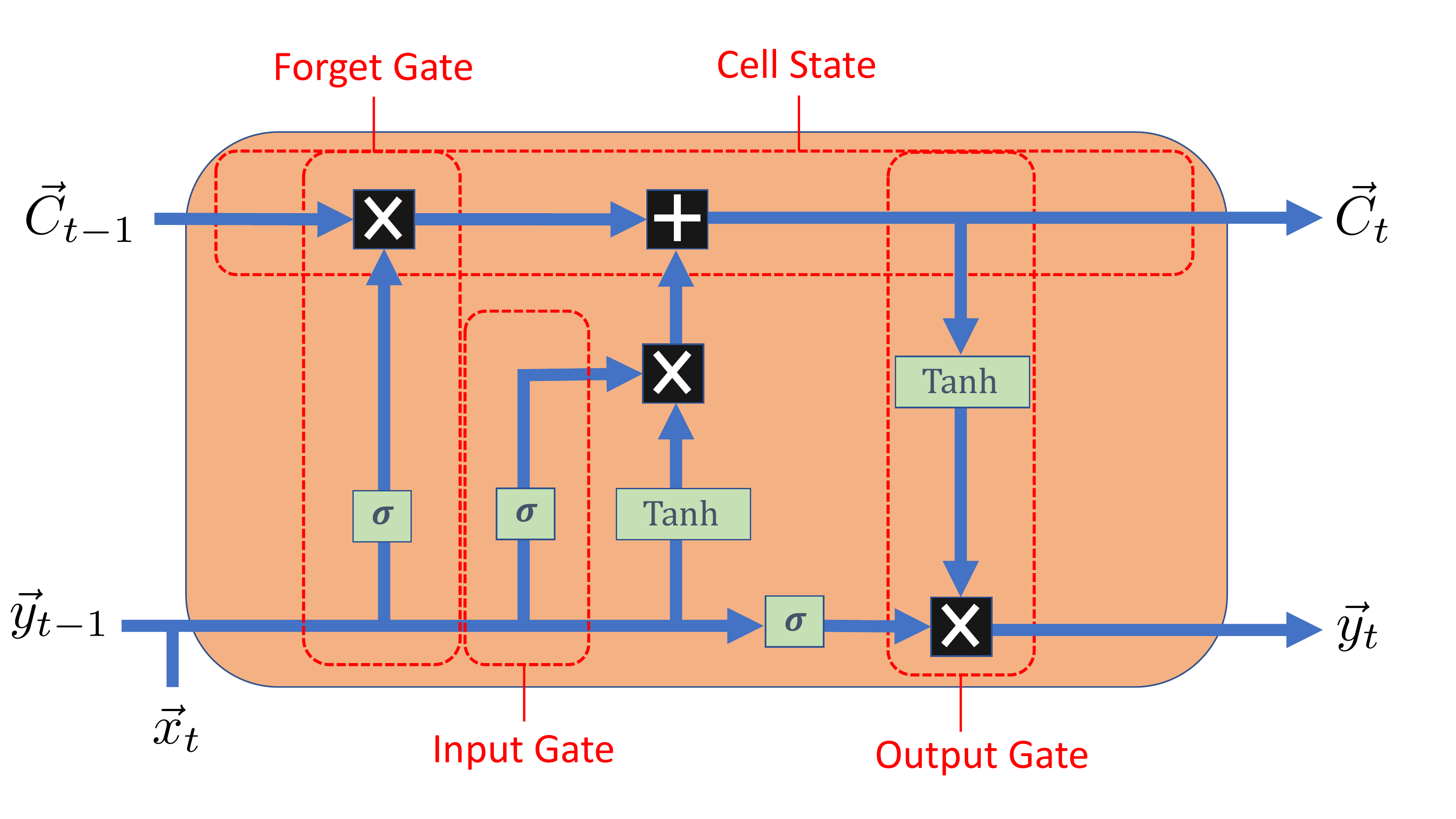}

\caption{Structure of the LSTM used in the prediction of the traffic flux.}

\label{fig:arquitectura}

\end{figure}

In Fig.\ref{fig:arquitectura} we describe the architecture of the Neural Network used for the analysis, the LSTM.  Such Neural Networks are very useful in forecasting problems, often surpassing traditional forecasting methods~\cite{siami2018comparison}. They are particularly powerful when dealing with complex datasets, with many points in a multidimensional space. This is indeed our case, as each time slot corresponds to many entries (location IDs) and their corresponding traffic flux measurement. 

The LSTM produces a set of  predictions at step $t$, which we label  ($\Vec{y}_t$), see Fig.~\ref{fig:arquitectura}. These  predictions  will be affected by the cell state (the long term memory that contains information kept by the Neural Network from all the previous steps, $\Vec{C}_{t-1}$), the short-term memory (the information of the label in the near previous steps, $\Vec{y}_{t-1}$) and all the features in the current step, $\Vec{x}_t$. When the LSTM calculates the prediction for $\Vec{y}_t$, it also decides which information should be retained and which to forget, thus updating the cell state $\Vec{C}_{t}$.

\begin{figure}[h]

\centering

\includegraphics[width = 0.8\textwidth]{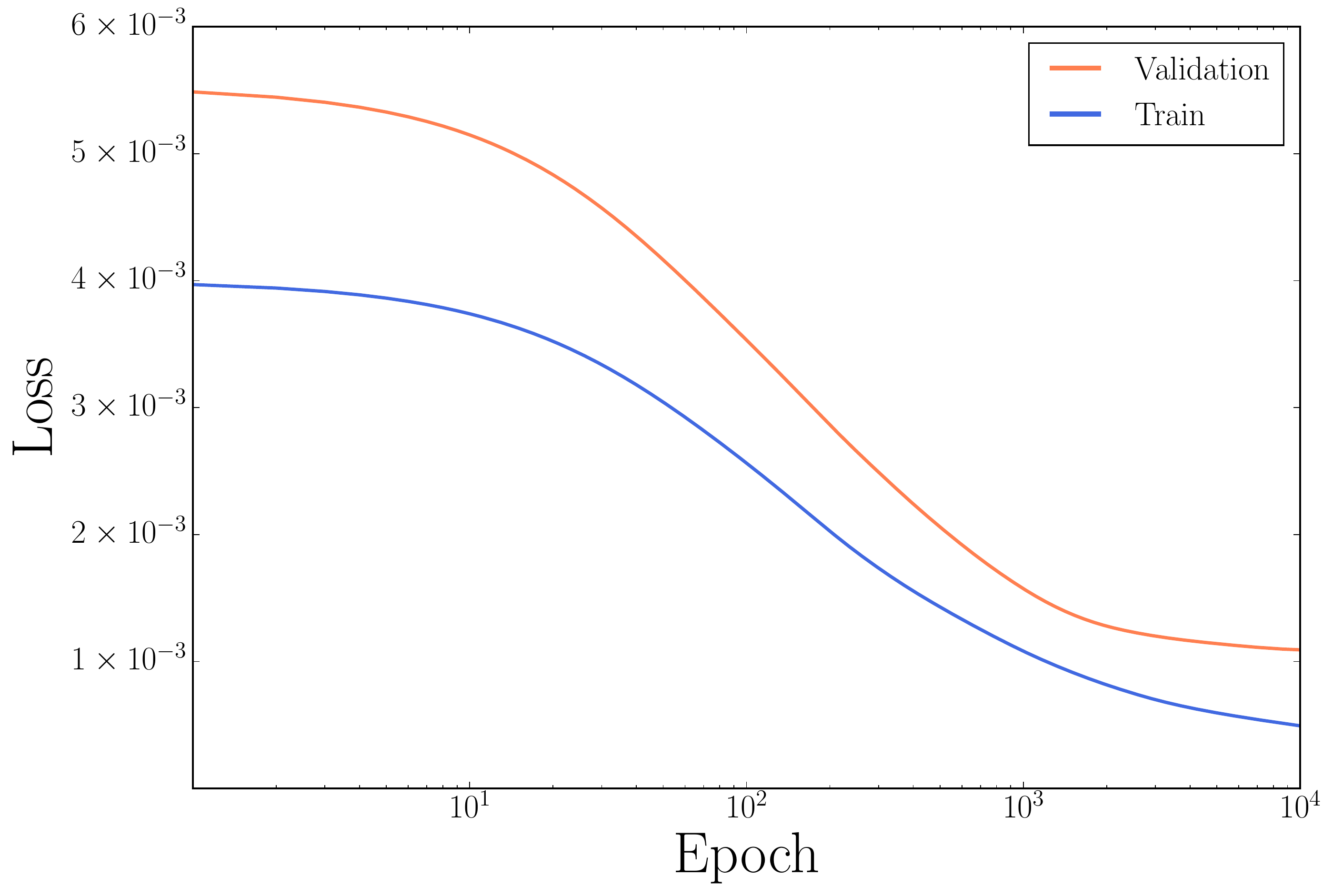}

\caption{Root mean square error of the neural network for the validation (orange) and training (blue) data sets per training epoch. The example chosen in the figure corresponds to the case where the Neural Network predicts 1h using the past 24h.}

\label{fig:loss}

\end{figure}

In all cases under study, we trained the LSTM during 11000 epochs. This number of epochs is  justified by analysing  Fig.~\ref{fig:loss}. In this figure, we plot the RMSE (Root-Mean-Squared Error) loss of the neural network per epoch: the loss decreases until about  10000 epochs, then starts to plateau, even on this scale that is logarithmic in the x-axis. The computational cost of training beyond this number of epochs is thus not justified: for instance, for 11000 epochs, the relative improvement in RMSE is negligible $|RMSE(10k) - RMSE(11k) | / RMSE(10k) < 10^{-5}$.

\subsection{Results}

The input to the learning algorithm is information on flux from all ID segments in 1h time steps. The LSTM is trained to predict the flux 1h or 2h ahead by looking back several hours.

Before we move on to show predictions, let us discuss whether the prediction target should be local (one segment) or global (all segments).  We find that focusing on the prediction of a particularly busy road, or asking for predictions for the whole network leads to very similar precision. Indeed, we chose the busy access road {\color{blue} Av. del Cid} discussed in the previous sections, to predict its traffic flux by analysing 24 h before and predicting 1h later. We then  ran two different algorithms, one  LSTM which predicted only the  {\color{blue} Av. del Cid} flux, and another LSTM predicting all  road segments, but both using all segments as input. The RMSE obtained in both cases was $RMSE = 320$ with a mean flux of $2200$ per hour.   Due to this result, in this paper, we only show the results of  predictions for this busy road segment, as it is computationally cheaper. However, all the analysis can be extrapolated to the case where we would predict all road segments of the city.

\subsubsection{Traffic prediction and uncertainty}

Once we have chosen a road segment that we are interested in, we can proceed to studying the stability and forecasting errors of our LSTM approach. For the rest of the analysis, we  forecast the traffic flux of {\color{blue} Av. del Cid} during the week between 02/02/2017 and 08/02/2017. 

\begin{figure}[h!]
\centering
\includegraphics[width = 1.00\textwidth]{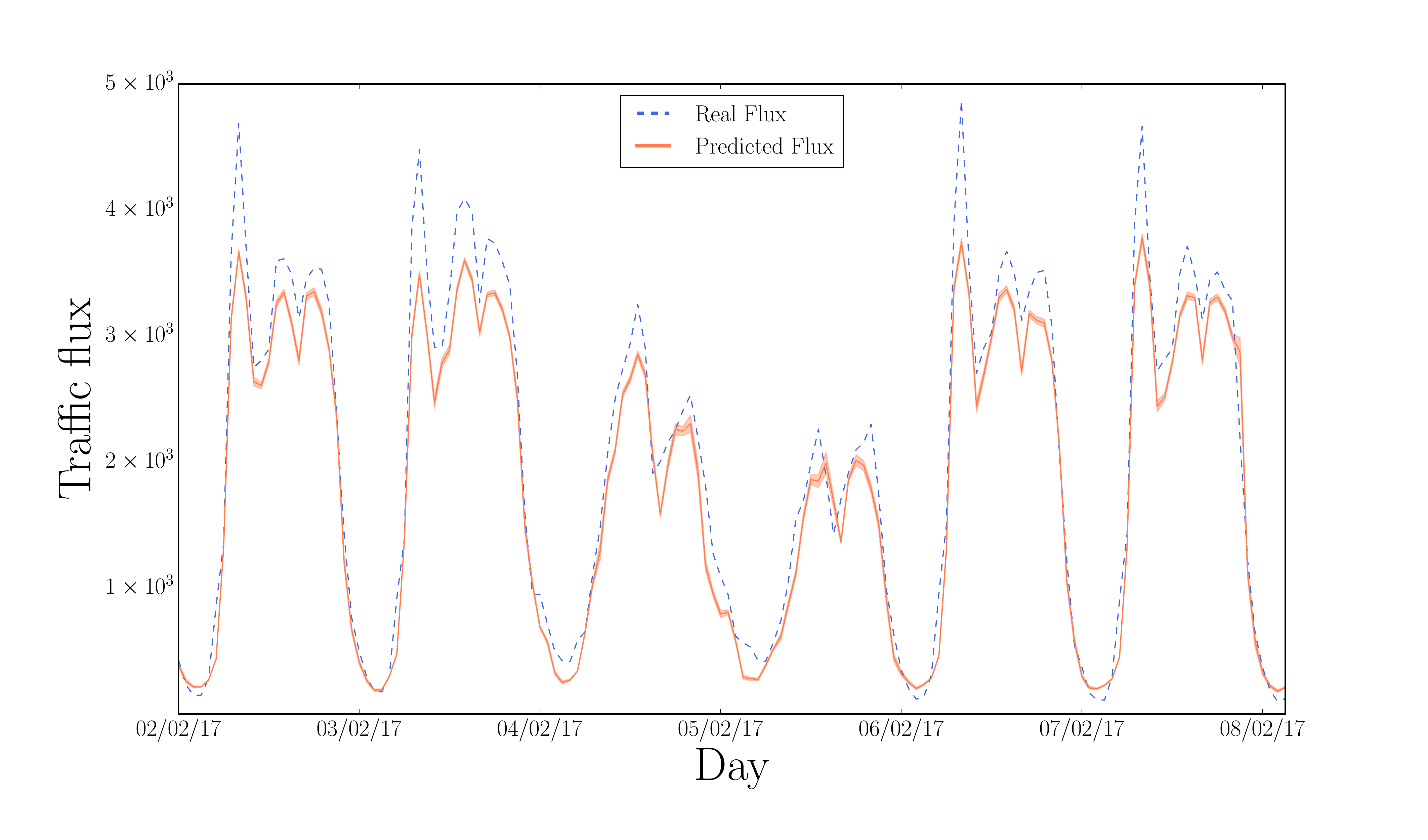}
\caption{Real flux (blue dashed line) and the mean prediction (orange line) for {\color{blue} Av. del Cid}  after training the neural network, with the 2016 data, ten different times. The shadow orange region represents the standard deviation of the mean prediction.}
\label{fig:bandas_error}
\end{figure}

In Fig.~\ref{fig:bandas_error} we show, on the one hand, the mean prediction (orange line) for the {\color{blue} Av. del Cid} after training the LSTM with the 2016 data ten different times. The orange shadow band is the standard deviation of the forecasting. In the plot, we show the prediction 1h ahead, calculated using the previous 24h.

In this figure, the error band is barely wide enough to be seen for most predicted points. More concretely, the error band is always between $0.5\%$ and  $10\%$ of the mean-predicted value, and this upper value of the relative error only occurs during times where the flux is near its (night-time) minimum.  
As we have asserted that the uncertainty in the prediction is small, in the next figures we will show comparisons using only the mean predictions. 

Another feature in this plot is that the LSTM tends to underpredict the flux during sharp increases. This result is expected, and is a typical behaviour of a  time-series predictor. The LSTM takes information about the events that have taken place in the previous hours, and one should expect a lower performance when predicting the flux peak in isolated hours. However, despite the expected error in the forecasting of the flux in those hours, we see that the Neural Network encapsulates the right  tendency, predicting  the time of a  high flux and an approximate value of the flux.

In forecasting problems, it is very common to find that the predicted curve shifts to the right compared to the real data. This would occur if the algorithm had not actually learnt real patterns in the data and, as consequence, the forecasting for the step $y_t$ would be essentially the same as the real value for the previous step $y_{t-1}$.  However, in this study, we do not observe any such displacement in the prediction curve, which shows that our LSTM has learned patterns from the data.

\subsubsection{The effect of local information in the prediction}

We now move on to discuss the impact of the training data on the prediction. As explained before, we have information on fluxes from thousands of sensorized segments in the city. Since we are looking to predict traffic for a single point in the city, the question arises of whether this coverage is really useful or whether one could get away with using less training data, for instance by restricting the inputs to local road segments.

It may be that a large coverage is needed to produce reliable predictions, if previous events in distant places of the city impact the prediction on the target segment. To explore this question we study the prediction for the traffic flux of {\color{blue} Av. del Cid} in three different cases, where we changed the coverage in road segments : using as input either {\it 1.)} only the location of the predicted segment (yellow line), {\it 2.)} the predicted segment and the nearby road segments (blue line) or, {\it 3.)} all the sensorized road segments (red line).  

\begin{figure}[h!]
\centering
\includegraphics[width = 1.00\textwidth]{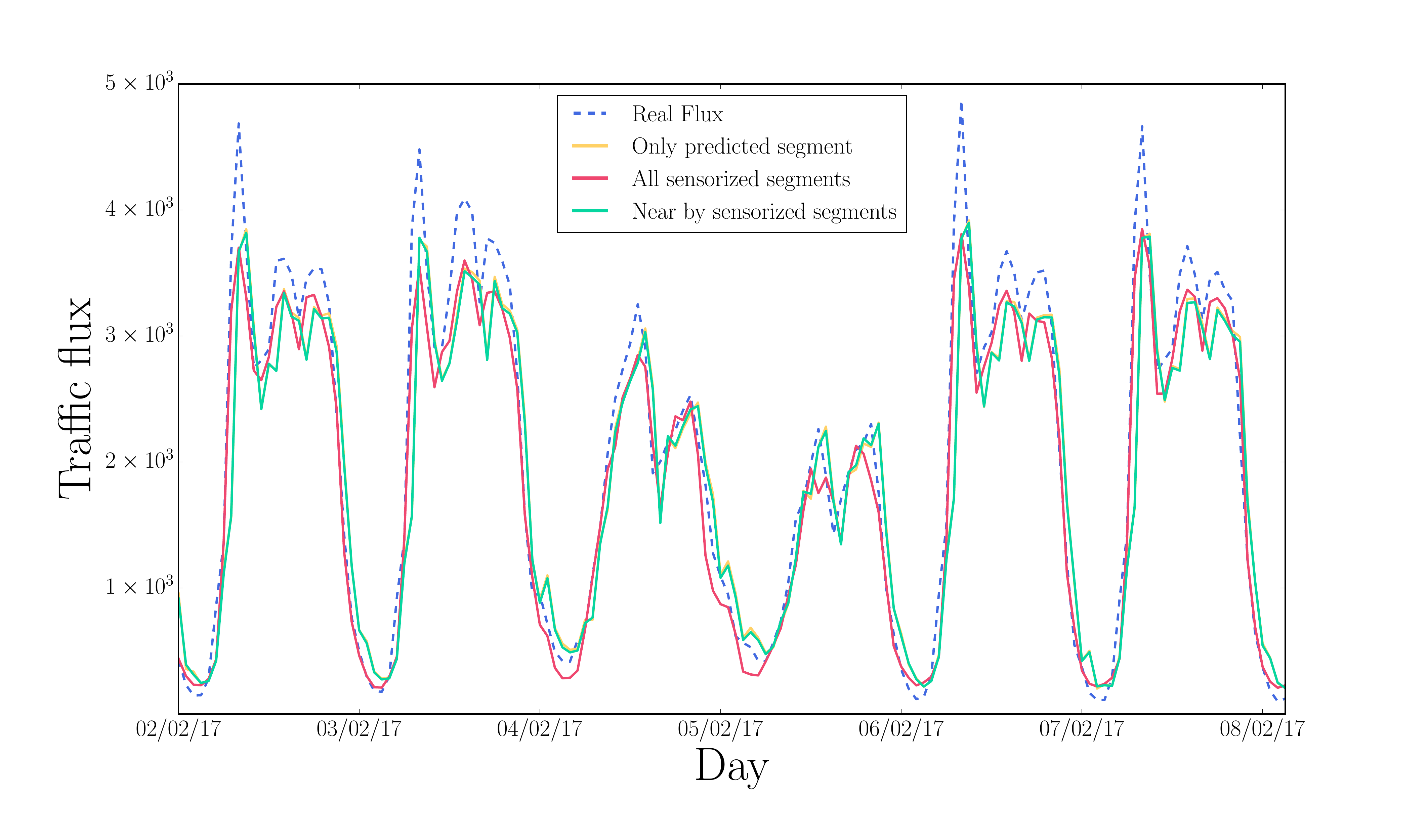}
\caption{Comparison between the three different training paradigms under study for the input: with all sensorized roads (pink line), roads near the segment being studied (green line) and only the output road (yellow line).}
\label{fig:tres_paradigmas}
\end{figure}

The results are shown in Fig.~\ref{fig:tres_paradigmas}. In all three cases, we predict 1h into the future, using the past 24h of data as an input (training and inference).

First, note that the three input cases produce predictions with high accuracy. The predictions using only the target road segment and using the nearby road segments exhibit  very similar accuracy, i.e. the yellow and green lines are practically indistinguishable practically identical. However, one can observe in these two lines a small shift to the right with respect to the dashed blue line (real data), an effect we discussed in the previous section and that indicates that the LSTM is not fully learning patterns of behaviour. 

This small shift is not present in the red line, indicating that knowledge of a wider set of road segments is adding valuable information. Indeed, the case where we use the input of all sensorized road segments is different. As we can see in the Figure, the prediction  is an excellent match to the real data except for the highest peaks. The shift observed in the other two cases has completely disappeared. Therefore, we conclude that having input/training data with a wide coverage of the traffic map is key to improving the predictions. 

\subsubsection{The effect of changing the length of future and past windows}

\begin{figure}[h!]
\centering
\includegraphics[width = 1.00\textwidth]{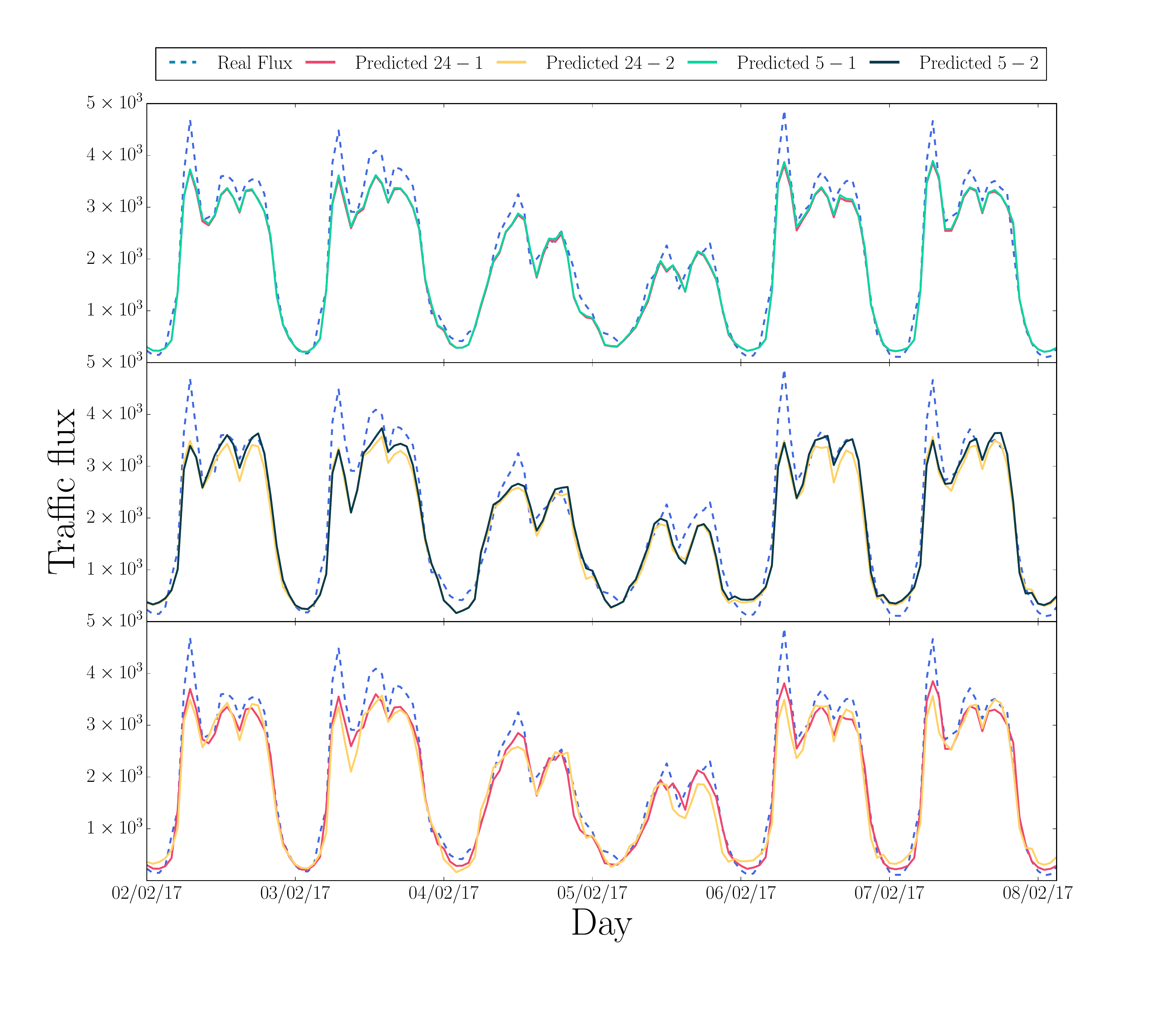}
\caption{Comparison of the forecasts using different time windows in the past and future. The first panel shows the prediction 1h into the future, using either 5h (green) or 24h (red) of past data ---the two curves are nearly indistinguishable. In the second panel, we compare predictions 2h ahead, using either 5h (black) or 24h (yellow) of past data ---a 5h back window yields better predictions. Finally, in the third panel, we represent the predictions using 24h of past data, predicting either 1h ahead (red) or 2h ahead (yellow) ---a 2h forward window does worse. In the three panels, the blue dashed line represents the real flux.}
\label{fig:comparacion_tiempos}
\end{figure}

Finally, we study how many hours of past data are needed for good predictions, and how many hours ahead we are able to predict well. In Fig.~\ref{fig:comparacion_tiempos} we show the results of using different time windows for training and predicting: using information from the past 24h to predict the next 1h (pink line) or 2h (yellow line) and, using the past 5h in order to predict 1h (clear blue line) and 2h (dark blue line) into the future.

\begin{figure}[h!]
\centering
\includegraphics[width = 1.00\textwidth]{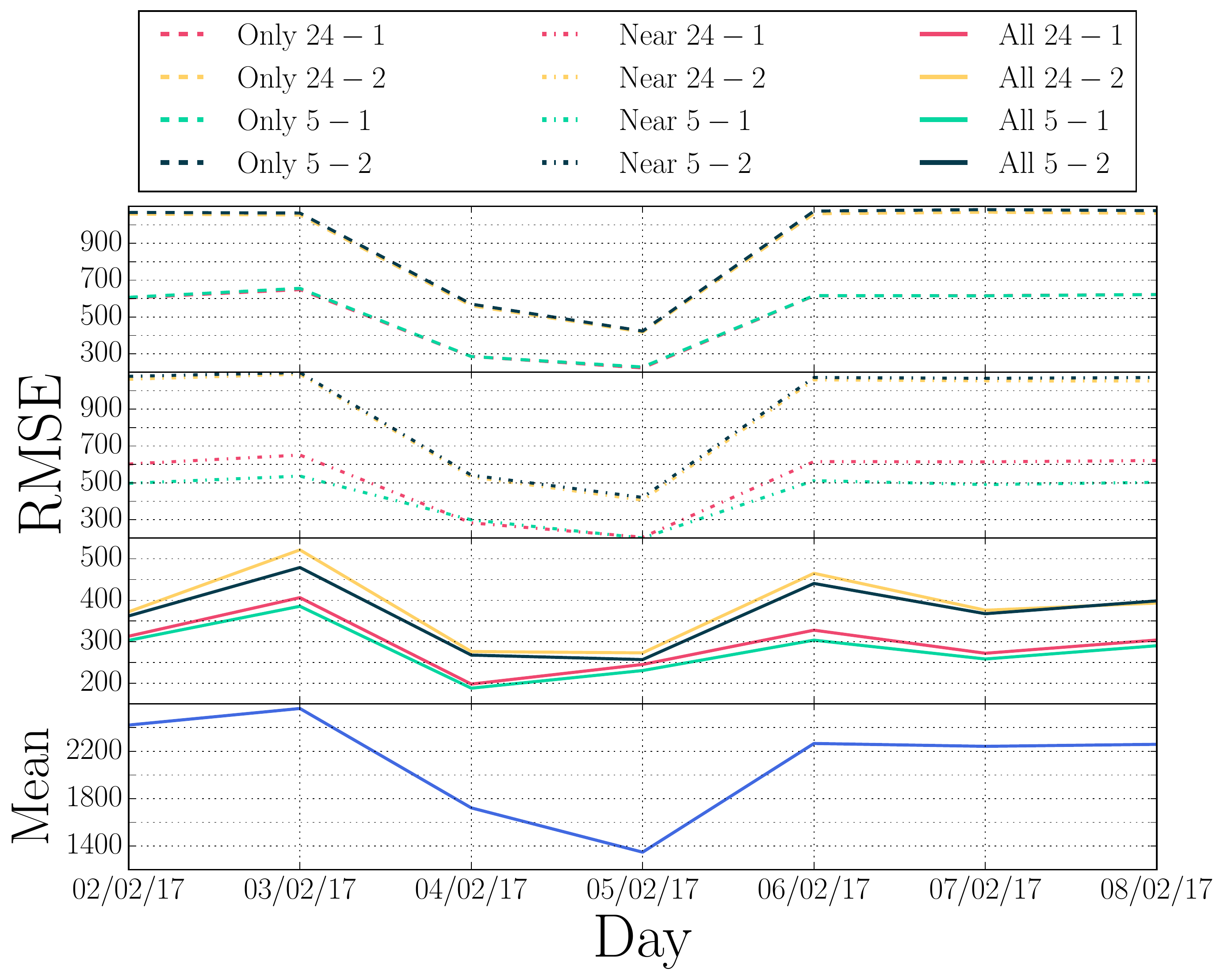}
\caption{Root mean square error (RMSE) per day in the studied week with the three studied input configurations for {\color{blue} Av. del Cid}: only past data form the segment to be predicted, past data from nearby road segments, and past data from all sensorized road segments. In each of the three first panels we show the four prediction cases: 24h or 5h of past data as input, and predictions 1h or 2h into the future. The last panel shows the mean of the traffic flux in the studied road segment per day for the chosen week.}
\label{fig:RMSE_comparacion}
\end{figure}

To show all the options and compare them, we divided the plot into three different panels in Fig.~\ref{fig:comparacion_tiempos}. 
Additionally, for completeness in Fig.~\ref{fig:RMSE_comparacion},  we plot the RMSE of the different cases studied in this paper. 

In the first panel of Fig.~\ref{fig:comparacion_tiempos}, we compare the actual flux (blue-dashed line) with the prediction for the next 1h, using 24h (pink) and 5h (clear blue) of past data. As one can see, including 24h of past data does very little to improve the predictions. In fact, it turns out that the error is slightly higher when we use 24h, see Fig.~\ref{fig:RMSE_comparacion}. In that Figure, we see that the distance between the RMSE using 24h or 5h is practically constant, and that the prediction using only 5h before is better. This observation makes sense if we consider that the prediction of the traffic flux, should only really be affected by the events that occur only a few hours before. The use of 24h time series, in this case, is not only computationally expensive, it also makes the Neural Network worse at extracting temporal patterns.

In the second panel of Fig.~\ref{fig:comparacion_tiempos} we plot the same as before, but predicting 2h instead of 1h ahead. The yellow and black line represent, respectively, the prediction using 24h and 5h of past data. As in the first panel, the best prediction is obtained using 5h of past data. However, unlike the 1h prediction cases were the improvement was small, here the 2h prediction is clearly better using 5h of past data instead of 24h. As in the first case, in the bottom panel of Fig.~\ref{fig:RMSE_comparacion} is clear that the RMSE of the  24h back - 2h forward case (yellow curve) is the highest.  

Finally, in the third panel of Fig.~\ref{fig:comparacion_tiempos}, we compare the prediction for the next 1h (pink) or 2h (yellow), using 24h of past data. As expected, the results are better when we attempt to predict fewer hours ahead.

The last figure that we show, Fig.~\ref{fig:RMSE_comparacion}, as mentioned previously, represents the different RMSEs in all cases being studied.

In the first panel (dashed lines), we can see the RMSEs for predictions using past data from the segment under consideration only, with the 4 combinations of duration for the past input data and for the forecast window represented as different colors. In that case, there is no difference between using the past 24h or 5h of data to train, and the curves are indistinguishable. On the other hand, we see that we are better at predicting 1h ahead rather than 2h ahead, as expected. Also, the lowest absolute errors are achieved on the weekend (4th and 5th of February), for which the mean traffic is also lower (see bottom panel in the Figure).

The second panel (in dash-dotted lines) shows the RMSEs for all cases, this time taking as input all road segments within 1km. This obviously increases the computing costs, yet the RMSE is very similar in all cases, except for one: predicting 1h into the future with 5h of past data. In that case, we get improved predictions by including nearby road segments rather than a single one: the LSTM is beginning to uncover the correlations between neighboring road segments and using this information to improve predictions. This only holds for the case with the short windows of time (5h into the past, 1h into the future).

In the third panel, we increase the computational cost again, using past data from all available sensorized road segments in order to predict our single segment of interest. In that case the LSTM is able to extract even more correlations between road segments, and we get a noticeable reduction in errors for all time windows (including when using 24h in past and predicting 2h into the future) compared to the top two panels. Yet, once again, we see that the best case is the one with the shortest time windows (5h backward, 1h forward): more past data seems to drown the relevant signal, making it harder for the LSTM to extract it, and predictions further into the future are always harder.

\section{Discussion and Outlook}

In this paper we have analyzed traffic data from the city of Valencia during the years 2016 and 2017. 

We first identified several temporal and spatial patterns. We observed three traffic peaks on   the  working days Monday to Thursday corresponding to work/study travel and lunchtime.  This workday pattern changed on Friday, signaling a variation in return times from work. We also found that the weekend days only display two peaks of traffic: midday and early evening. 
Looking at the spatial distribution of this flux, we found interesting variations in incoming and outgoing traffic behaviour, especially  in the most congested roads. All these findings could be used to design better intervention plans. 

We then moved to an analysis based on predicting traffic flux one to two hours ahead of time. For this task, we trained  several Neural Networks and explored their performance under various situations and explored the impact of network coverage and past and future time windows for predictions. 

Overall we found an excellent capacity of predicting the flux, even at rush hours. For example, our analysis of the busiest access road in Valencia produced predictions with errors ranging from 15\% during weekdays to 20\% during weekends, see Fig.~\ref{fig:RMSE_comparacion}. In other words, in the busiest roads at the busiest moments during the day, the accuracy of the algorithm's prediction was close to 85\%.

We also noticed that, even though our model was not aware of the geographical relations between road segments, and even though our data contained lots of disconnected road segments (fewer than 20\% of all road segments), accuracy improved a lot by including data from road segments farther than 1km. This means that the LSTM does learn correlations between road segments, even thought the 
connections between them are not made explicit in the data, and most road segments are missing anyway. On the other hand, we noted that there was no gain to be had from including past data beyond 5h in order to predict the next 1h. This might change if we included data from more than 24h, perhaps using the attention mechanism.

This study could motivate the development of a real-time tool to predict the traffic in the city of Valencia. Also, the Neural Network could be used to test different intervention techniques, e.g. by simulating the effect of road closures or traffic light controls,  and allowing the algorithm we trained with real data to produce predictions under these interventions.   

\section*{Funding Statement}

The work of MGF is supported by the {\it Margarita Salas} postdoctoral fellowship from the Ministerio de Educaci\'on UP2021-044.  The research of VS is supported by the Generalitat
Valenciana PROMETEO/2021/083 and the Ministerio de Ciencia e
Innovacion PID2020-113644GB-I00. The work of JFU and EGL is supported by the Agència Valenciana de la Innovació (AVI) INNEST/2021/263.



\bibliographystyle{elsarticle-num} 
\bibliography{cas-refs}





\end{document}